\documentclass[]{bytedance_seed}
\usepackage[T1]{fontenc}
\usepackage[toc,page,header]{appendix}

\usepackage{wrapfig}
\usepackage{multirow}
\usepackage{makecell}
\usepackage{xcolor}
\usepackage{colortbl}

\usepackage{amsmath,amsfonts,bm}

\usepackage{xspace}

\newcommand{\markedblanklines}[2][\textbullet]{
  \par
  \begingroup
  \setlength{\parskip}{0pt}
  \setlength{\parindent}{0pt}
  \count0=0
  \loop
    \ifnum\count0<#2
      #1\hspace{1em}\rule{0pt}{\baselineskip}\par
      \advance\count0 by 1
  \repeat
  \endgroup
}

\def\eqref#1{equation~\ref{#1}}

\def\1{\bm{1}}

\DeclareMathAlphabet{\mathsfit}{\encodingdefault}{\sfdefault}{m}{sl}
\SetMathAlphabet{\mathsfit}{bold}{\encodingdefault}{\sfdefault}{bx}{n}

\definecolor{colorfirst}{rgb}{.866,.945, 0.831}
\definecolor{colorsecond}{rgb}{1, 0.98, 0.83}
\definecolor{colorthird}{rgb}{0.76, 0.87, 0.92}
\definecolor{colorcite}{rgb}{0.212, 0.490, 0.741}

\usepackage{cleveref}
\crefname{figure}{Fig.}{Figs.}
\crefname{table}{Tab.}{Tabs.}
\crefname{equation}{Eq.}{Eqs.}
\crefname{section}{Sec.}{Secs.}

\title{HD-VGGT: \\ High-Resolution Visual Geometry Transformer}

\author[1,2, *,\dagger]{Tianrun Chen}
\author[1,*]{Yuanqi Hu}
\author[3,*]{Yidong Han}
\author[1, *]{Hanjie Xu}
\author[4, *]{Deyi Ji}
\author[4, *]{Qi Zhu}
\author[5]{Chunan Yu}
\author[1]{Xin Zhang}
\author[1]{Cheng Chen}
\author[1]{Chaotao Ding}
\author[3]{Ying Zang}
\author[6]{Xuanfu Li} 
\author[6]{~~~~ Jin Ma}
\author[7]{Lanyun Zhu}

\affiliation[1]{KOKONI 3D, Moxin Technology}
\affiliation[2]{Zhejiang University}
\affiliation[3]{Huzhou University}
\affiliation[4]{Univeristy of Science and Technology
of China}
\affiliation[5]{Nanjing University of Science and Technology}
\affiliation[6]{Huawei}
\affiliation[7]{Tongji University}

\contribution[\dagger]{Project Lead}
\contribution[*]{Equal Contribution}

\abstract{
High-resolution imagery is essential for accurate 3D reconstruction, as many geometric details only emerge at fine spatial scales. Recent feed-forward approaches, such as the Visual Geometry Grounded Transformer (VGGT), have demonstrated the ability to infer scene geometry from large collections of images in a single forward pass. However, scaling these models to high-resolution inputs remains challenging: the number of tokens in transformer architectures grows rapidly with both image resolution and the number of views, leading to prohibitive computational and memory costs. Moreover, we observe that visually ambiguous regions, such as repetitive patterns, weak textures, or specular surfaces, often produce unstable feature tokens that degrade geometric inference, especially at higher resolutions. We introduce HD-VGGT, a dual-branch architecture for efficient and robust high-resolution 3D reconstruction. A low-resolution branch predicts a coarse, globally consistent geometry, while a high-resolution branch refines details via a learned feature upsampling module. To handle unstable tokens, we propose Feature Modulation, which suppresses unreliable features early in the transformer. HD-VGGT leverages high-resolution images and supervision without full-resolution transformer costs, achieving state-of-the-art reconstruction quality.
}

\correspondence{Tianrun Chen, Lanyun Zhu}

\begin{document}

\maketitle

\footnotetext[1]{We thank Jianyuan Wang for his insightful discussions. We acknowledge the support from Hisilicon, the ZJU Kunpeng \& Ascend Center of Excellence, and the Dream Set Off - Kunpeng \& Ascend Seed Program.}

\section{Introduction}

Recovering the three-dimensional structure of the world from images is a central goal of computer vision. Accurate 3D reconstruction enables machines to perceive and interact with complex environments, powering applications from robotics and autonomous navigation to augmented reality and digital content creation. Classical pipelines built on Structure-from-Motion (SfM) and Multi-View Stereo (MVS) \cite{schonberger2016sfm,schonberger2016mvs} have made remarkable progress over the past decades, but they rely on intricate multi-stage optimization procedures that are computationally expensive and often brittle in challenging real-world conditions.

Recent advances in deep learning are beginning to reshape this landscape. Instead of carefully orchestrated optimization pipelines, feed-forward models aim to infer scene geometry directly from images. Among them, the Visual Geometry Grounded Transformer (VGGT) \cite{vggt} represents a major step forward. By leveraging global self-attention, VGGT jointly processes hundreds of views and predicts camera poses together with dense scene geometry in a single forward pass. This paradigm shift, from iterative optimization to large-scale neural inference, opens a promising path toward faster and more scalable 3D reconstruction systems.

Despite this progress, achieving high-resolution reconstruction remains a fundamental challenge. Fine geometric structures, such as thin objects, sharp edges, and subtle surface variations, are often only visible at high spatial resolutions. At the same time, increasing the number of input views typically provides stronger geometric constraints and improves reconstruction accuracy. An ideal reconstruction system should therefore be able to process many high-resolution images efficiently, fully exploiting the rich visual information available in real-world image collections.

However, scaling VGGT to this regime is inherently difficult. In transformer-based architectures, the number of tokens grows rapidly with both the number of input images and their spatial resolution. As either factor increases, computation and memory requirements quickly become prohibitive. In practice, this forces existing feed-forward reconstruction models to operate under strict limits on image resolution and view count, preventing them from fully leveraging the information necessary for high-fidelity reconstruction.

Beyond this scalability bottleneck, we further observe a phenomenon that limits reconstruction accuracy. In visually ambiguous regions, such as repetitive patterns, weak textures, or specular surfaces, the early layers of transformer-based models often produce statistically unstable feature tokens. These tokens respond inconsistently across views and propagate noise through subsequent layers, ultimately degrading geometric reasoning. As image resolution increases, such unstable regions become more prominent, making them a critical obstacle for high-precision reconstruction.

In this work, we introduce HD-VGGT, a new architecture designed for efficient and robust high-resolution feed-forward 3D reconstruction. Our approach adopts a memory-efficient dual-branch design. A low-resolution branch first estimates a globally consistent coarse 3D representation using the original VGGT backbone. Rather than scaling the entire network to higher resolutions, which would incur prohibitive computational cost, we design a dedicated feature upsampling module that forms the core of a high-resolution refinement branch. This module fuses coarse 3D features with high-resolution image features, enabling the model to leverage both the geometric structure already inferred and the rich visual information present in high-resolution inputs. In doing so, the network can benefit from high-resolution supervision while avoiding the quadratic complexity explosion of full-resolution transformer processing.

To further improve robustness, we introduce a Feature Modulation mechanism that detects unstable feature tokens in early transformer layers and attenuates their influence during feature propagation. By reducing the impact of unreliable features, this mechanism produces cleaner geometric signals and enables more stable multi-view reasoning.

Taken together, these ideas lead to a reconstruction framework that is both scalable and robust. HD-VGGT enables feed-forward models to operate at substantially higher resolutions while maintaining computational efficiency, and it improves geometric reliability in challenging visual conditions. We believe this direction opens new opportunities for large-scale, high-fidelity 3D reconstruction, bringing feed-forward approaches closer to the level of detail and robustness required for real-world applications.

\section{Related Work}

\subsection{Feed-Forward 3D Reconstruction}
The field of modern visual perception has experienced a fundamental evolution toward scalable, data-driven representation learning and unified modeling strategies. This evolution has driven substantial improvements across a broad range of computer vision applications, from semantic interpretation to geometric inference \cite{arnold2022map,zhu2025skysense,zhu2024ibd,fastvggt,reizenstein2021common,zhu2025replay,wang2025pi,zhu2025cpcf,ji2026view,vggt}.
Feed-forward 3D reconstruction has emerged as a dominant paradigm, offering an efficient alternative to classical optimization-based SfM~\cite{schonberger2016sfm} and MVS~\cite{schonberger2016mvs} pipelines. The seminal DUSt3R~\cite{dust3r} introduced the concept of direct point map regression from image pairs, a direction quickly advanced by numerous follow-up works. These include architectural refinements for scalability like  Fast3R~\cite{yang2025fast3r}, as well as methods for handling long video sequences like Spann3R~\cite{wang2024spann3r}. The introduction of global attention in VGGT~\cite{vggt} marked a significant step towards more robust multi-view understanding, with variants like $\pi^3$~\cite{wang2025pi} addressing issues such as reference-view bias. Concurrently, methods like Dens3R~\cite{fang2025dens3r} and CUT3R~\cite{wang2025cut3r} have explored denser geometric predictions and mixed static/dynamic data training. Other notable contributions include FLARE~\cite{zhang2025flare} and AnySplat~\cite{jiang2025anysplat}, which bridge foundation models with 3D Gaussian Splatting for novel view synthesis. Despite this rapid progress, a common limitation persists: these models are predominantly designed for and evaluated on standard-resolution imagery. The challenge of efficiently scaling these powerful architectures to high-resolution inputs, without compromising accuracy, remains a critical and underexplored frontier.

\subsection{High-Resolution Deep Learning}
The challenge of processing high-resolution imagery is not unique to 3D vision. The broader computer vision community has explored numerous strategies to mitigate the computational burden of high-resolution inputs in deep neural networks \cite{liu2021swin, ji2024discrete, skysenseo, ji2023ultra, ji2022structural, popen, ji2025structural, zhu2025not, pptformer, stlnet, gpwformer, zhu2024llafs, cagcn, zhu2025llafs++}. These include the development of efficient transformer variants with linear or near-linear complexity attention mechanisms \cite{efficient_transformers}, the use of multi-scale architectures that process images at different resolutions \cite{multiscale_arch}, and patch-based processing techniques that divide large images into smaller, manageable chunks. Our work builds upon the insights from this field, but instead of modifying the core attention mechanism, we adopt a hierarchical, dual-branch approach that separates coarse, global reasoning from fine-grained, local refinement, offering a more direct solution to the problem of high-resolution 3D reconstruction.

\section{Method}

\begin{figure*}[ht]
  \centering
  \includegraphics[width=\linewidth]{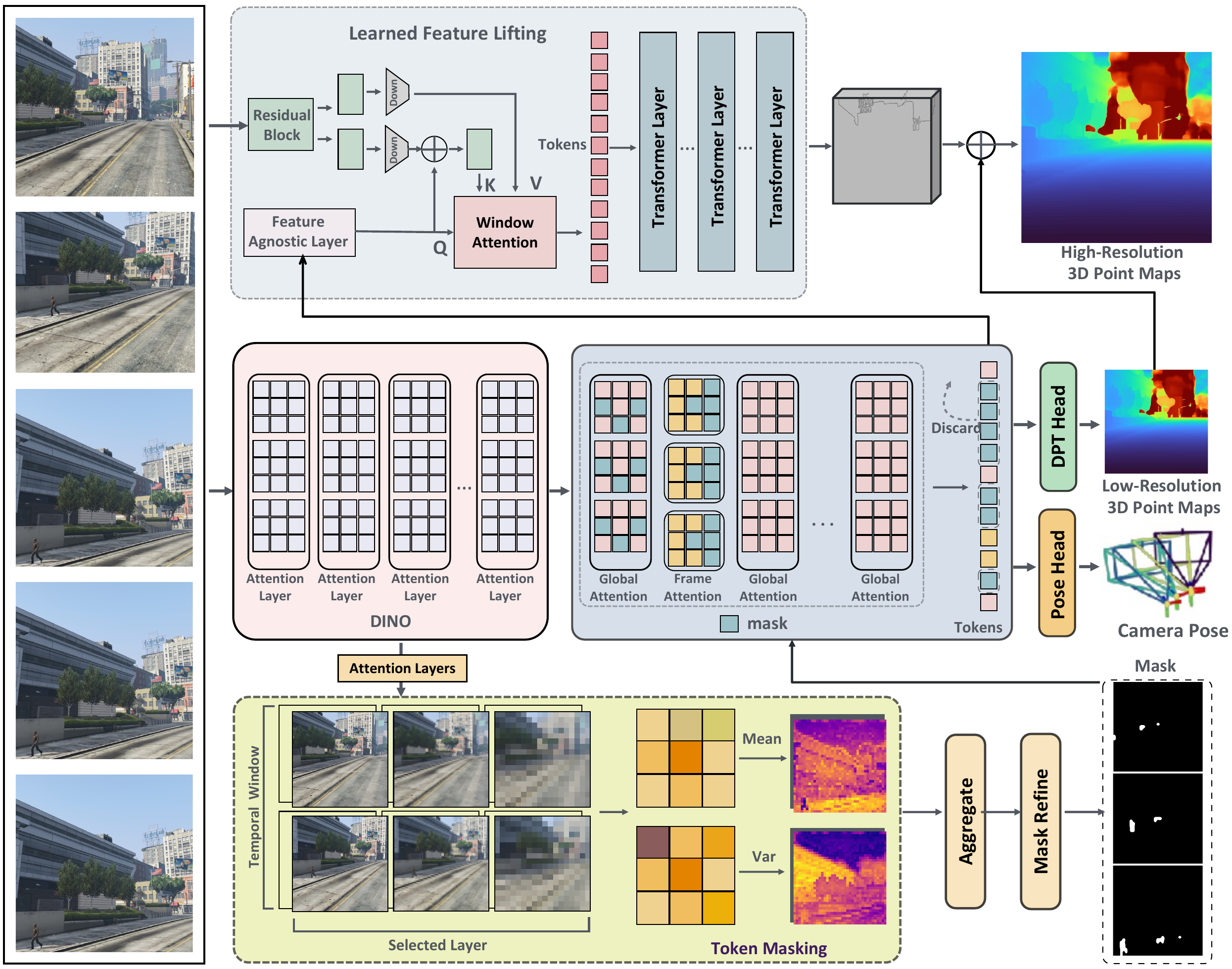} 
  \caption{Overview of the HD-VGGT architecture. A low-resolution branch first computes a coarse 3D feature volume. A high-resolution branch then uses a learned feature upsampler guided by the high-resolution image to produce detailed features, which are processed by a lightweight refiner transformer to yield the final high-fidelity output.}
  \label{fig:arch}
\end{figure*}

\subsection{Preliminaries: Revisiting VGGT Bottleneck}

The standard VGGT architecture processes a set of $N$ images $\{I_i \in \mathbb{R}^{H_0 \times W_0 \times 3}\}_{i=1}^N$. Each image is passed through a Vision Transformer (ViT) backbone, such as DINOv2 \cite{dinov2}, to produce a sequence of $K_0$ patch tokens $F_i \in \mathbb{R}^{K_0 \times C}$. The core of VGGT is a deep transformer that applies global self-attention across the concatenated tokens from all $N$ views, a total of $N \cdot K_0$ tokens. The computational and memory complexity of this global attention operation is $O((N \cdot K_0)^2 \cdot C)$, which is quadratic in the total number of tokens.

This quadratic scaling presents a significant barrier to high-resolution processing. If we aim to double the input resolution to $2H_0 \times 2W_0$, the number of tokens per image, $K$, increases by a factor of four (i.e., $K \approx 4K_0$). Consequently, the complexity of the global attention mechanism explodes by a factor of $4^2 = 16$, rendering high-resolution reconstruction computationally infeasible with current hardware. This fundamental limitation motivates our hierarchical approach.

\subsection{HD-VGGT: A Hierarchical Dual-Branch Architecture}

To circumvent the quadratic complexity bottleneck while harnessing the detail present in high-resolution imagery, we propose a dual-branch architecture, as illustrated in Figure~\ref{fig:arch}. The design philosophy is to decouple the task of establishing a global, coarse geometric scaffold from the task of refining fine-grained local details.

\subsubsection{Low-Resolution Branch}

This branch is responsible for efficient, global 3D inference. It takes as input a set of images $\{I_i^{LR}\}_{i=1}^N$ downsampled to a standard resolution (e.g., $518 \times 518$ pixels). These images are processed by a standard VGGT backbone, which consists of a DINOv2 feature extractor and a deep transformer network $\mathcal{T}_{coarse}$:

\begin{equation}
    \{F_i^{coarse}, g^{coarse}\} = \mathcal{T}_{coarse}(\{I_i^{LR}\}_{i=1}^N)
\end{equation}

This branch outputs a set of coarse, low-resolution 3D-aware feature maps $\{F_i^{coarse} \in \mathbb{R}^{h \times w \times c}\}$ and a set of initial camera parameters $g^{coarse}$. By operating at a low resolution, this branch efficiently establishes the overall scene structure and camera poses, providing a robust scaffold for the subsequent refinement stage.

\subsubsection{High-Resolution Branch} This branch is designed for detail refinement and operates in two stages:

\textit{1) Learned Feature Upsampling.} The transition from the coarse geometric scaffold $F^{coarse}$ to a high-resolution representation presents a critical challenge. A naive interpolation, such as bilinear or nearest upsampling, can be formulated as a linear combination of neighboring feature vectors:
\begin{equation}
    F^{hr}(p) = \sum_{q \in \mathcal{N}(p')} (1 - |p'_x - q_x|)(1 - |p'_y - q_y|) F^{coarse}(q)
\end{equation}where $p$ is a coordinate in the high-resolution grid, $p' = p/s$ is its corresponding scaled coordinate in the low-resolution grid (with $s$ being the downsampling factor), and $\mathcal{N}(p')$ are the four nearest integer-coordinate neighbors.This operation, while computationally trivial, acts as a low-pass filter, inevitably leading to the loss of high-frequency geometric and textural details, which are essential for high-fidelity reconstruction. 

To overcome this information-theoretic bottleneck, we posit that the high-resolution image $I^{HR}$ itself contains the necessary high-frequency cues to guide the upsampling process. We therefore employ a learned, guidance-based feature upsampling module, denoted $\mathcal{U}$, inspired by recent advances in this area \cite{anyup}. The core idea is to model the upsampling process as a conditional generation problem, where the high-resolution feature map $F^{hr}$ is generated conditioned on both the coarse feature map $F^{coarse}$ and the high-resolution guidance image $I^{HR}$.

We formulate this as learning a mapping $\mathcal{U}: (\mathbb{R}^{h \times w \times c}, \mathbb{R}^{H \times W \times 3}) \rightarrow \mathbb{R}^{H \times W \times c}$. The module first extracts high-frequency details from the guidance image using a shallow convolutional network $\phi_{guidance}$. Concurrently, the coarse features are upsampled to the target resolution using a simple interpolation followed by a convolutional refinement network $\phi_{feat}$. The two streams are then fused and processed by a final convolutional network $\phi_{fuse}$:
\begin{align}
    F_{guide} &= \phi_{guidance}(I^{HR}) \label{eq:guidance_feat} \\
    F_{interp} &= \phi_{feat}(\text{Interpolate}(F^{coarse})) \label{eq:interp_feat} \\
    F^{hr} &= \phi_{fuse}(\text{concat}(F_{guide}, F_{interp})) \label{eq:upsample_final}
\end{align}
This formulation allows the network to learn to inject high-frequency, spatially-precise details from the guidance image into the geometrically sound but coarse feature map, effectively reversing the information loss from downsampling and producing a feature representation that is both geometrically consistent and rich in fine-grained detail.

\textit{2) High-Resolution Refiner.} The upsampled feature maps $\{F_i^{hr}\}_{i=1}^N$ are then processed by a lightweight refiner transformer, $\mathcal{T}_{refine}$. This network is significantly shallower than the backbone transformer (e.g., 6 layers vs. 24 layers) and can employ more efficient, localized attention mechanisms, as the global geometry has already been resolved. Its purpose is to refine the high-resolution features, enforce multi-view consistency at a fine scale, and regress the final, high-fidelity outputs:

\begin{equation}
    \{D^{final}, g^{final}\} = \mathcal{T}_{refine}(\{F_i^{hr}\}_{i=1}^N)
\end{equation}

This hierarchical decomposition allows HD-VGGT to achieve a high-resolution output without ever needing to compute global self-attention over a massive number of high-resolution tokens, thus maintaining computational tractability.

\subsection{Feature Modulation for Robust Geometric Inference}

The pursuit of high-fidelity geometric reconstruction at elevated resolutions necessitates a profound understanding of the underlying feature manifold. An ideal multi-view feature representation should constitute a low-dimensional, isometrically embedded Riemannian manifold, where geodesic distances correspond to semantic and geometric dissimilarities. However, real-world scenes invariably contain what we term geometric singularities, such as  regions of repetitive texture, textureless surfaces, or specular highlights. These singularities disrupt the smooth structure of the feature manifold, introducing high-frequency noise and topological defects that corrupt the geometric signal. From an information-theoretic perspective, these regions represent points of high Kolmogorov complexity and Shannon entropy, fundamentally challenging the assumption of cross-view feature consistency.

Inspired by the analysis of dynamic scenes \cite{vggt4d}, we posit that these static geometric singularities are, from a statistical standpoint, isomorphic to dynamic events. Both phenomena manifest as a breakdown in spatio-temporal feature consistency, leading to a high-variance stochastic process when observing a 3D point's features across multiple views. To address this, we introduce a principled, training-free framework for identifying and suppressing these anomalous feature tokens by analyzing the higher-order statistics of the feature space.

\subsubsection{Manifold Anomaly Detection via Kernelized Gramian Statistics}

The standard attention mechanism, which operates on first-order feature vectors, is insufficient to capture the complex, non-linear distortions introduced by geometric singularities. We therefore elevate our analysis to a Reproducing Kernel Hilbert Space (RKHS)  defined by the feature embeddings themselves. The geometry of the data in this space is captured by the Gram matrix. For a given layer $l$ and a pair of views $(t, s)$, we compute the cross-view Gramians:
\begin{equation}
    G^{QQ}_{l,t,s} = \frac{Q_{l,t} Q_{l,s}^T}{\sqrt{c}}, \quad G^{KK}_{l,t,s} = \frac{K_{l,t} K_{l,s}^T}{\sqrt{c}}
\end{equation}
These Gramians serve as kernel matrices that encode the complete second-order geometry of the feature manifold between views.

We model the sequence of these Gramians across a temporal window of neighboring views $\mathcal{W}(t)$ as a matrix-valued stochastic process. The first and second moments of this process provide a powerful description of the manifold's stability. We define two statistical operators, the temporal expectation $\mathbb{E}_{\mathcal{W}}[\cdot]$ and the temporal variance $\mathbb{V}_{\mathcal{W}}[\cdot]$, to estimate these moments over a set of layers $\mathcal{L}_{i,j}$:
\begin{align}
    S^{X}_{i-j} &= \mathbb{E}_{\mathcal{W}} [ \mathbb{E}_{\mathcal{L}_{i,j}} [ G^{X}_{l,t,s} ] ] \label{eq:mean_gram} \\
    V^{X}_{i-j} &= \mathbb{V}_{\mathcal{W}} [ \mathbb{E}_{\mathcal{L}_{i,j}} [ G^{X}_{l,t,s} ] ] \label{eq:var_gram}
\end{align}
Here, $S^{X}_{i-j}$ represents the stationary component (the expected geometric structure) of the manifold, while $V^{X}_{i-j}$ quantifies its volatility or “information-theoretic surprise”—a direct measure of anomalous behavior.

Following the hierarchical Bayesian inference framework proposed in \cite{vggt4d}, we construct a posterior anomaly saliency map, $\mathcal{S}_{anomaly}$, by fusing multi-scale priors derived from different network depths:
\begin{equation}
    \mathcal{S}_{anomaly} = \omega_{shallow} \odot \omega_{middle} \odot \omega_{deep}
\end{equation}
where the priors are defined as:
\begin{align}
    \omega_{shallow} &= (\mathbf{1} - S^{KK}_{shallow}) \odot V^{QK}_{shallow} \label{eq:w_shallow} \\
    \omega_{middle} &= \mathbf{1} - S^{QQ}_{middle} \label{eq:w_middle} \\
    \omega_{deep} &= (\mathbf{1} - V^{QQ}_{deep}) \odot S^{QQ}_{deep} \label{eq:w_deep}
\end{align}
In this formulation, $\omega_{shallow}$ acts as a high-frequency semantic anomaly prior, $\omega_{middle}$ provides a mid-frequency geometric instability prior, and $\omega_{deep}$ serves as a low-frequency spatial regularization prior. Their product yields a robust posterior estimate of the anomaly probability for each token.

\subsubsection{Manifold Regularization via Projection Gradient Flow}

The initial anomaly map, $\mathcal{M}_{initial} = [\mathcal{S}_{anomaly} > \alpha]$, represents a coarse segmentation of the manifold's singularities. To refine this, we perform a manifold regularization step guided by the gradient flow of the projection error. We consider the 3D point cloud as a discrete sampling of the scene manifold, and define a vector field on this manifold based on reprojection errors. The total aggregated gradient for a point, $\nabla_{\mathcal{A}}$, is a function of the geometric residual $r_{d,i}$ and photometric residual $r_{c,i}$:
\begin{align}
    \nabla_{\mathcal{A}} &= \frac{1}{N} \sum_i^N \| w_i r_{d,i} \nabla r_{d,i} \| + \lambda \frac{1}{N} \sum_i^N \| w_i r_{c,i} \| \label{eq:agg_grad}
\end{align}
where $w_i$ masks the computation to the initially stable regions. The magnitude of this gradient represents the local distortion of the manifold. By thresholding this distortion metric, we obtain a refined anomaly mask $\mathcal{M}_{refined}$ that more accurately delineates the boundaries of the geometric singularities.

\subsubsection{Information-Gated Attentional Suppression}

Finally, we must prevent the propagation of uncertainty from these identified anomalous regions. We conceptualize this as an information-gating problem. A naive, full-depth suppression would be equivalent to completely closing the information channel, leading to out-of-distribution behavior. Instead, we implement a targeted early-stage information-gating mechanism. We apply the refined mask $\mathcal{M}_{refined}$ only to the shallow layers $\mathcal{L}_{shallow}$, where the feature encoding is most susceptible to corruption. The gating is achieved by nullifying the Key vectors of anomalous tokens:
\begin{equation}
    K'_{p,l} = (1 - \mathcal{M}_{refined, p}) \cdot K_{p,l}, \quad \forall l \in \mathcal{L}_{shallow}
\end{equation}
This operation effectively closes the information gate for corrupted data streams at their source, preventing error amplification in deeper layers. This ensures that the subsequent geometric reasoning and high-resolution refinement stages operate on a purified, high-integrity information substrate, which is a critical prerequisite for achieving state-of-the-art reconstruction fidelity.

\section{Experiments}

\subsection{Experimental Setup}

\noindent\textbf{Datasets.} We evaluate HD-VGGT across a diverse set of standard benchmarks to demonstrate its versatility. For camera pose estimation, we use RealEstate10K \cite{zhou2018realestate10k} for static scenes, Sintel \cite{sintel} and TUM-dynamics \cite{sturm2012tum} for dynamic environments, and CO3Dv2 \cite{co3d} for object-centric analysis. Point map reconstruction is evaluated on 7-Scenes \cite{7scene}, NRGBD \cite{azinovic2022nrgbd}, DTU \cite{jensen2014dtu}, and high-resolution MVS-SYNTH \cite{huang2018mvsynth}. For monocular depth estimation, we test on ScanNet \cite{dai2017scannet} and NYUv2 \cite{silberman2012nyu} under various pose conditions, including scenarios without ground-truth camera priors.

\noindent\textbf{Evaluation Metrics.} We follow task-specific protocols for a fair comparison. For camera pose, we report AUC at $10^\circ$ and $30^\circ$, RRA, RTA, and trajectory metrics including ATE and RPE. Point map quality is measured by Accuracy and Completeness (Mean and Median in cm). Depth estimation is quantified using Abs Rel and $\delta_1$ accuracy, with all predictions aligned to the ground-truth scale via least-squares fitting.

\noindent\textbf{Baselines.} We compare our method against a broad spectrum of state-of-the-art feed-forward frameworks. These include geometry transformers such as VGGT, $\pi^3$, and WorldMirror, as well as DUSt3R-paradigm models including DUSt3R, MASt3R, Fast3R, and CUT3R. We also evaluate against other influential works like FLARE, DA3, and MonST3R. To isolate the impact of our proposed components, we include an ablation model HD-VGGT (w/o FM) which disables the Feature Modulation mechanism.

\noindent\textbf{Implementation Details.} The model is trained for 2 weeks on a cluster of 16 Ascend 910C within CloudMatrix384 supernode. We adopt a similar training strategy to VGGT, including data scheduling, optimizer settings, and loss formulations.

\begin{figure*}[ht]
  \centering
  \includegraphics[width=0.95\linewidth]{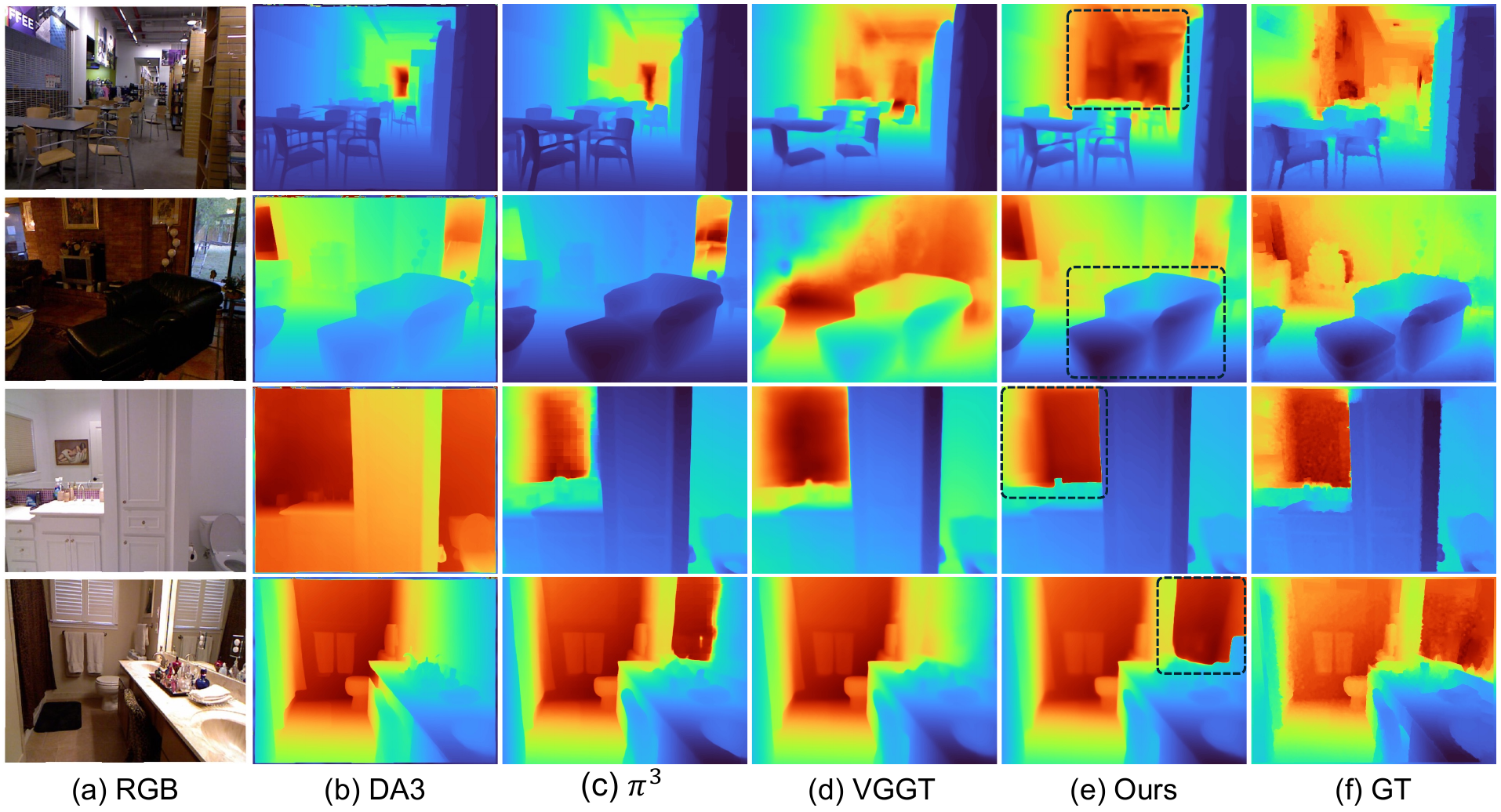}
  \caption{Qualitative comparison of monocular depth estimation. Our method produces more accurate and detailed results, which are indicated by dotted boxes}
  \label{fig:vis_depth}
\end{figure*}

\begin{figure*}[ht]
  \centering
  \includegraphics[width=0.95\linewidth]{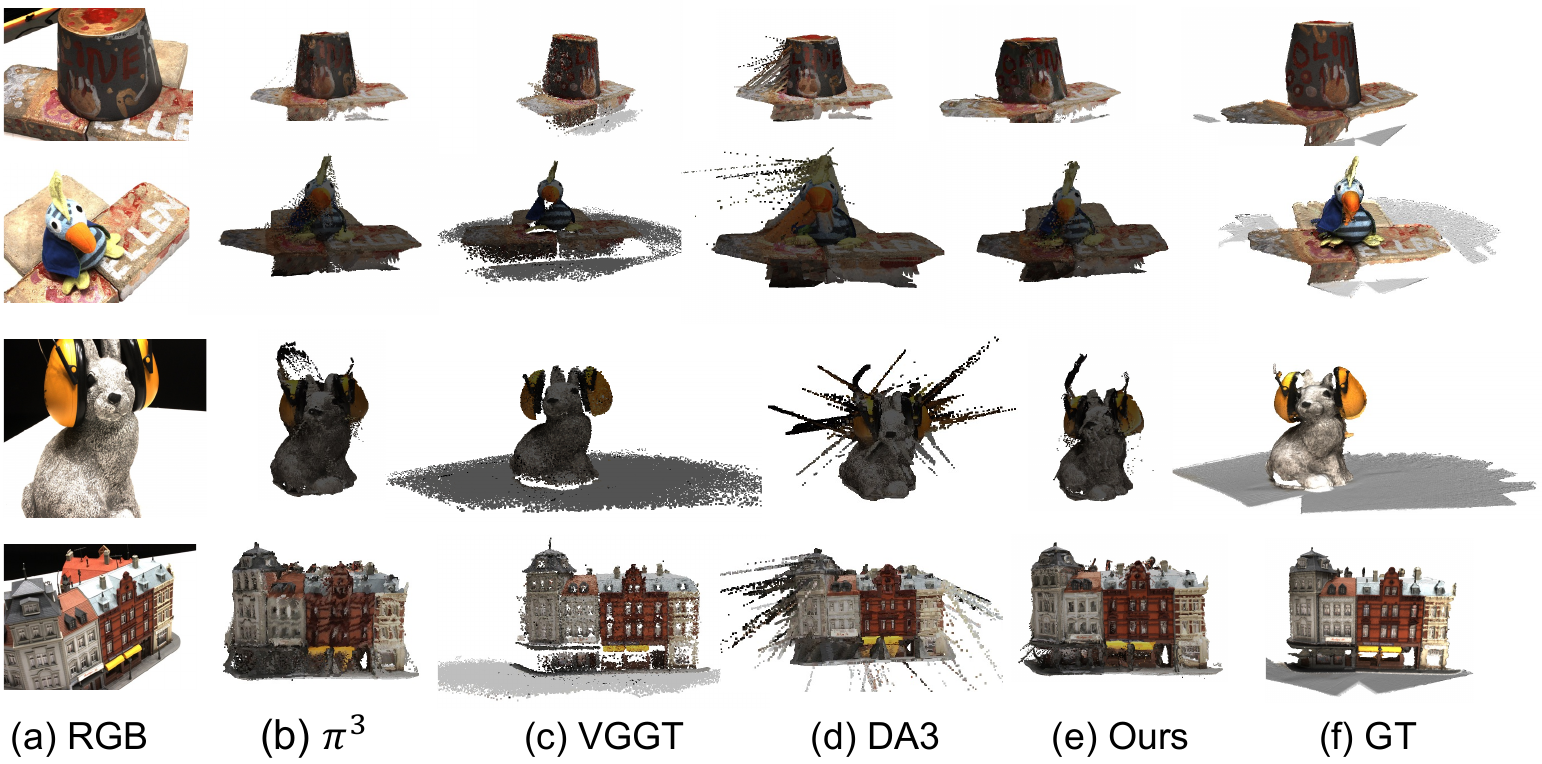} 
  \caption{Qualitative comparison of Point map reconstruction.}
  \label{fig:vis_pos}
\end{figure*}

\begin{table*}[t]
\centering
\caption{Camera pose estimation on RealEstate10K, Sintel, and TUM-dynamics. Our method demonstrates robust and accurate pose estimation.}
\resizebox{1\textwidth}{!}{
\begin{tabular}{@{}lccccccccc@{}}
\toprule
\multirow{2}{*}{Method} & \multicolumn{3}{c}{RealEstate10K (mixed, static)} & \multicolumn{3}{c}{Sintel (outdoor, dynamic)} & \multicolumn{3}{c}{TUM-dynamics (indoor, dynamic)} \\
\cmidrule(lr){2-4} \cmidrule(lr){5-7} \cmidrule(lr){8-10}
 & RRA@30$\uparrow$ & RTA@30$\uparrow$ & AUC@30$\uparrow$ & ATE$\downarrow$ & RPE$_t$$\downarrow$ & RPE$_r$$\downarrow$ & ATE$\downarrow$ & RPE$_t$$\downarrow$ & RPE$_r$$\downarrow$ \\
\midrule
Fast3R~\cite{yang2025fast3r} & 99.05 & 81.86 & 61.68 & 0.371 & 0.298 & 13.75 & 0.090 & 0.101 & 1.425 \\
CUT3R~\cite{wang2025cut3r} & 99.82 & 95.10 & 81.47 & 0.217 & 0.070 & 0.636 & 0.047 & 0.015 & 0.451 \\
FLARE~\cite{zhang2025flare} & 99.69 & 95.23 & 80.01 & 0.207 & 0.090 & 3.015 & 0.026 & 0.013 & 0.475 \\
VGGT~\cite{vggt} & 99.97 & 93.13 & 77.62 & 0.167 & 0.062 & 0.491 & 0.012 & 0.010 & 0.312 \\
$\pi^3$~\cite{wang2025pi} & 99.99 & 95.62 & 85.90 & 0.074 & 0.040 & 0.282 & 0.014 & 0.009 & 0.312 \\
WorldMirror~\cite{liu2025worldmirror} & 99.99 & 95.81 & 86.28 & 0.096 & 0.058 & 0.490 & 0.010 & 0.009 & 0.297 \\
\midrule
\textbf{HD-VGGT (Ours)} & \textbf{99.99} & \textbf{96.15} & \textbf{87.01} & \textbf{0.071} & \textbf{0.038} & \textbf{0.275} & \textbf{0.009} & \textbf{0.008} & \textbf{0.281} \\
\bottomrule
\end{tabular}
}
\label{tab:camera}
\end{table*}

\begin{table}[t]
\centering
\caption{Comparison of camera pose estimation on CO3Dv2. Metrics are Pose AUC $\uparrow$ at $10^\circ$ and $30^\circ$ thresholds.}
\label{tab:co3dv2_pose}
\scalebox{1}{\begin{tabular}{@{}lccc@{}}
\toprule
Method & Pose AUC $\uparrow$ (@10) & Pose AUC $\uparrow$ (@30) \\
\midrule
DUSt3R \cite{dust3r} & 70.1 & 76.7 \\
MASt3R \cite{leroy2024mast3r} & 78.2 & 81.8 \\
VGGT \cite{vggt} & 83.3 & 88.2 \\
\midrule
\textbf{HD-VGGT (Ours)} & \textbf{86.5} & \textbf{90.4} \\
\bottomrule
\end{tabular}}
\end{table}

\begin{table*}[t]
\centering
\caption{Point map reconstruction on 7-Scenes, NRGBD, and DTU. Our HD-VGGT sets a new state-of-the-art across all benchmarks.}
\resizebox{1\textwidth}{!}{
\begin{tabular}{@{}lcccccccccccc@{}}
\toprule
& \multicolumn{4}{c}{7-Scenes (scene)} & \multicolumn{4}{c}{NRGBD (scene)} & \multicolumn{4}{c}{DTU (object)} \\
\cmidrule(lr){2-5} \cmidrule(lr){6-9} \cmidrule(lr){10-13}
Method & \multicolumn{2}{c}{Acc. (cm) $\downarrow$} & \multicolumn{2}{c}{Comp. (cm) $\downarrow$} & \multicolumn{2}{c}{Acc. (cm) $\downarrow$} & \multicolumn{2}{c}{Comp. (cm) $\downarrow$} & \multicolumn{2}{c}{Acc. (cm) $\downarrow$} & \multicolumn{2}{c}{Comp. (cm) $\downarrow$} \\
& Mean & Med. & Mean & Med. & Mean & Med. & Mean & Med. & Mean & Med. & Mean & Med. \\
\midrule
Fast3R~\cite{yang2025fast3r} & 9.6 & 6.5 & 14.5 & 9.3 & 13.5 & 9.1 & 16.3 & 10.4 & 334.0 & 191.9 & 292.9 & 112.5 \\
CUT3R~\cite{wang2025cut3r} & 9.4 & 5.1 & 10.1 & 5.0 & 10.4 & 4.1 & 7.9 & 3.1 & 474.2 & 260.0 & 340.0 & 131.6 \\
FLARE~\cite{zhang2025flare} & 8.5 & 5.8 & 14.2 & 10.4 & 5.3 & 2.4 & 5.1 & 2.5 & 254.1 & 146.8 & 317.4 & 142.0 \\
VGGT~\cite{vggt} & 4.6 & 2.6 & 5.7 & 3.4 & 5.1 & 2.9 & 6.6 & 3.8 & 133.8 & 77.9 & 189.6 & 99.2 \\
$\pi^3$~\cite{wang2025pi} & 4.8 & 2.8 & 7.2 & 4.7 & 2.6 & 1.5 & 2.8 & 1.4 & 119.8 & 64.6 & 184.9 & 60.7 \\
WorldMirror~\cite{liu2025worldmirror} & 4.3 & 2.6 & 4.9 & 2.8 & 4.1 & 2.0 & 4.5 & 1.9 & 101.7 & 56.4 & 178.0 & 69.0 \\
\midrule
\textbf{HD-VGGT (Ours)} & \textbf{3.9} & \textbf{2.1} & \textbf{4.5} & \textbf{2.5} & \textbf{2.4} & \textbf{1.3} & \textbf{2.6} & \textbf{1.2} & \textbf{95.3} & \textbf{51.2} & \textbf{165.4} & \textbf{61.1} \\
\bottomrule
\end{tabular}
}
\label{tab:pointmap}
\end{table*}

\begin{table}[h]
\centering
\caption{Monocular depth estimation on ScanNet and NYUv2.}
\begin{tabular}{@{}lcccc@{}}
\toprule
\multirow{2}{*}{Method} & \multicolumn{2}{c}{ScanNet} & \multicolumn{2}{c}{NYUv2} \\
\cmidrule(lr){2-3} \cmidrule(lr){4-5}
 & AbsRel $\downarrow$ & $\delta_1$ $\uparrow$ & AbsRel $\downarrow$ & $\delta_1$ $\uparrow$ \\
\midrule
DUSt3R~\cite{dust3r} & 0.081 & 0.909 & 0.143 & 0.814 \\
MASt3R~\cite{leroy2024mast3r} & 0.110 & 0.865 & 0.115 & 0.848 \\
VGGT~\cite{vggt} & 0.056 & 0.951 & 0.062 & 0.969 \\
$\pi^3$~\cite{wang2025pi} & 0.054 & 0.956 & 0.038 & 0.986 \\
WorldMirror~\cite{liu2025worldmirror} & 0.052 & 0.957 & 0.063 & 0.968 \\
\midrule
\textbf{HD-VGGT (Ours)} & \textbf{0.049} & \textbf{0.961} & \textbf{0.035} & \textbf{0.988} \\
\bottomrule
\end{tabular}
\label{tab:depth}
\end{table}

\begin{table}[t]
\centering
\caption{Monocular Depth estimation on ScanNet under different pose conditions. Metrics are Abs Rel $\downarrow$ evaluated with and without ground-truth (GT) poses.}
\label{tab:scannet_depth}
\begin{tabular}{@{}lcc@{}}
\toprule
Method & \begin{tabular}[c]{@{}c@{}}Depth Abs Rel $\downarrow$\\ (w/ GT Pose)\end{tabular}  & \begin{tabular}[c]{@{}c@{}}Depth Abs Rel $\downarrow$\\ (w/o GT Pose)\end{tabular} \\
\midrule
DUSt3R \cite{dust3r} & 0.145 & 0.182 \\
MASt3R \cite{leroy2024mast3r} & 0.131 & 0.165 \\
VGGT \cite{vggt} & 0.119 & 0.140 \\
\midrule
\textbf{HD-VGGT (Ours)} & \textbf{0.102} & \textbf{0.121} \\
\bottomrule
\end{tabular}
\end{table}

\subsection{Main Results}

We conduct a comprehensive evaluation of HD-VGGT against state-of-the-art methods on a wide range of standard benchmarks, following the protocols established by recent works like WorldMirror~\cite{liu2025worldmirror}. We focus on three core tasks: Camera Pose Estimation, Point Map Reconstruction, and Monocular Depth Estimation.

\subsubsection{Camera Pose Estimation}

We evaluate the camera pose estimation performance of HD-VGGT across four diverse datasets: RealEstate10K (static), Sintel (dynamic outdoor), TUM-dynamics (dynamic indoor), and CO3Dv2 (object-centric). As summarized in Table~\ref{tab:camera} and Table~\ref{tab:co3dv2_pose}, our model consistently outperforms state-of-the-art methods such as $\pi^3$, WorldMirror, and MASt3R across all scenarios. On RealEstate10K, HD-VGGT achieves a superior AUC@30 of $87.01\%$, while on the more challenging CO3Dv2 benchmark, it significantly improves the high-precision Pose AUC@10 to $86.5\%$, demonstrating exceptional accuracy in both scene-level and object-level alignment. In dynamic environments, our method reduces the Absolute Trajectory Error (ATE) to $0.071$ on Sintel and $0.009$ on TUM-dynamics, outperforming previous feed-forward frameworks. These results indicate that the high-resolution dual-branch architecture, enhanced by feature modulation, effectively mitigates ambiguities caused by transient dynamics and specular reflections, leading to more robust and precise global pose optimization.

\subsubsection{Point Map Reconstruction}

For the point map reconstruction task, HD-VGGT is evaluated on the 7-Scenes, NRGBD, and DTU benchmarks, where it establishes a new state-of-the-art by significantly outperforming all baseline models across all metrics. As shown in Table \ref{tab:pointmap}, our method achieves superior precision with an accuracy of $3.9$ cm on 7-Scenes and $2.4$ cm on NRGBD, while also reaching a new benchmark on the challenging DTU dataset with a mean accuracy of $95.3$ cm. These improvements demonstrate that the high-resolution input allows the model to capture significantly finer geometric details and produce more precise point clouds compared to standard-resolution feed-forward architectures.

\subsubsection{Monocular Depth Estimation}

We evaluate the monocular depth estimation performance of HD-VGGT on the ScanNet and NYUv2 datasets. As presented in Table~\ref{tab:depth} and Table~\ref{tab:scannet_depth}, our method achieves the best performance across all metrics, yielding an AbsRel of $0.049$ on ScanNet and $0.035$ on NYUv2. Compared to the base VGGT and recent SOTA methods like $\pi^3$ and WorldMirror, HD-VGGT shows a significant improvement in depth accuracy (AbsRel) and pixel reliability ($\delta_1$). Furthermore, we assess the robustness of our depth estimation on ScanNet under different pose conditions. Even in the more challenging setting without ground-truth poses (w/o GT Pose), HD-VGGT maintains high precision with an AbsRel of $0.121$, outperforming competitive frameworks such as MASt3R and DUSt3R. These results underscore the efficacy of our dual-branch architecture, which leverages high-resolution features to recover precise local geometry while maintaining strong global structural consistency, even in complex indoor environments where camera priors are unavailable.

\subsection{Qualitative Analysis}

We present a visual comparison of monocular depth estimation results in Figure \ref{fig:vis_depth} to further highlight the advantages of our high-resolution dual-branch architecture. While state-of-the-art methods provide reasonable global layouts, they often suffer from over-smoothing and "bleeding" artifacts around object boundaries. 
In contrast, HD-VGGT yields depth maps with superior sharpness and structural fidelity. Specifically, in complex indoor scenes, our model successfully preserves the geometry of fine-grained objects, such as lamp poles and chair legs, which typically disappear in low-resolution feature maps. Moreover, the surfaces of walls and floors in our results are more planar and consistent, demonstrating that the High-Resolution Refiner effectively utilizes detailed visual cues to resolve depth ambiguities that occur at coarser scales.

We also provide a visual comparison of 3D point map reconstructions in Figure \ref{fig:vis_pos}. We observe that while the baseline VGGT can recover the general layout of the scenes, they frequently fail to capture complex or thin-structured objects. 
In contrast, HD-VGGT yields significantly more complete and dense reconstructions. By effectively modulating high-resolution features, our model suppresses geometric noise while preserving fine-grained spatial details. These results qualitatively confirm that HD-VGGT achieves a superior balance between global structural consistency and local geometric precision, even in scenes with high structural complexity.

\section{Conclusion and Discussion}

In this work, we presented HD-VGGT, a new architecture for efficient and robust high-resolution feed-forward 3D reconstruction. By introducing a dual-branch design with a dedicated feature upsampling module, our method enables the model to exploit high-resolution imagery and supervision without scaling the entire transformer backbone. In addition, the proposed Feature Modulation mechanism improves robustness by suppressing unstable feature tokens that arise in visually ambiguous regions. Together, these components allow HD-VGGT to produce more accurate and detailed reconstructions while maintaining computational efficiency.

Our results demonstrate that scaling feed-forward reconstruction models to higher resolutions requires not only architectural efficiency but also improved feature stability. We hope that the ideas presented in this work—particularly cross-resolution geometric refinement and feature-level robustness mechanisms—can inspire future research toward more scalable and reliable neural 3D reconstruction systems.

Looking forward, we believe that further integrating geometric reasoning, large-scale image collections, and high-resolution neural representations will continue to push the limits of feed-forward 3D reconstruction, bringing these systems closer to practical deployment in real-world applications.

\clearpage

\bibliographystyle{unsrt}
\bibliography{main2}

\clearpage

\end{document}